\documentclass{article}

\usepackage{microtype}
\usepackage{graphicx}
\usepackage{subcaption}  
\usepackage{booktabs} 

\usepackage{bm}  

\usepackage{hyperref}

\usepackage[preprint]{icml2023}

\usepackage{amsmath}
\usepackage{amssymb}
\usepackage{mathtools}
\usepackage{amsthm}

\usepackage[capitalize,noabbrev]{cleveref}

\theoremstyle{plain}

\theoremstyle{definition}

\theoremstyle{remark}

\newcommand{\res}[1]{$#1{\mkern-2mu\times\mkern-2mu}#1$}
\newcommand{\cider}{CIDEr}
\newcommand{\ngram}{$n$-gram}
\newcommand{\reinforce}{\textsc{Reinforce}}
\def\average{\mathop{\rm mean}}%
\DeclareMathOperator*{\E}{\mathbb{E}}

\usepackage[textsize=tiny]{todonotes}

\usepackage{array}
\newcommand{\PreserveBackslash}[1]{\let\temp=\\#1\let\\=\temp}
\newcolumntype{C}[1]{>{\PreserveBackslash\centering}p{#1}}
\newcolumntype{R}[1]{>{\PreserveBackslash\raggedleft}p{#1}}
\newcolumntype{L}[1]{>{\PreserveBackslash\raggedright}p{#1}}

\begin{document}

\twocolumn[
\icmltitle{Tuning computer vision models with task rewards}



\icmlsetsymbol{equal}{*}

\begin{icmlauthorlist}
\icmlauthor{Andr\'e Susano Pinto}{equal,goog}
\icmlauthor{Alexander Kolesnikov}{equal,goog}
\\
\icmlauthor{Yuge Shi}{goog,intern}
\icmlauthor{Lucas Beyer}{goog}
\icmlauthor{Xiaohua Zhai}{goog}
\end{icmlauthorlist}

\icmlaffiliation{goog}{Google Research, Brain Team Z\"urich}
\icmlaffiliation{intern}{Work done during internship at Google Research, while being a PhD student at the University of Oxford}

\icmlcorrespondingauthor{Andr\'e Susano Pinto}{andresp@google.com}
\icmlcorrespondingauthor{Alexander Kolesnikov}{akolesnikov@google.com}

\icmlkeywords{Machine Learning, ICML}

\vskip 0.3in
]



\printAffiliationsAndNotice{
\textsuperscript{*}Shared first authorship and leadership.
} 

\begin{abstract}
Misalignment between model predictions and intended usage can be detrimental for the deployment of computer vision models.
The issue is exacerbated when the task involves complex structured outputs, as it becomes harder to design procedures which address this misalignment.
In natural language processing, this is often addressed using reinforcement learning techniques that align models with a task reward.
We adopt this approach and show its surprising effectiveness across multiple computer vision tasks, such as object detection, panoptic segmentation, colorization and image captioning.
We believe this approach has the potential to be widely useful for better aligning models with a diverse range of computer vision tasks.
\end{abstract}

\section{Introduction}

\newlength\picheight
\setlength\picheight{0.85in}
\newcommand{\taskexample}[3]{
      \centering
      \parbox[t]{2mm}{\rotatebox[origin=c]{90}{\tiny{\,\,\,\,Input}}}
      \begin{tabular}{@{}c@{}}\includegraphics[height=\picheight]{#1}\end{tabular}
      {\LARGE$|$}
      \hspace{0.5em}
      \parbox[t]{2mm}{\rotatebox[origin=c]{90}{\tiny{\,\,\,\,Before}}}
      \begin{tabular}{@{}c@{}}\includegraphics[height=\picheight]{#2}\end{tabular}
      \hspace{0.3em}
      \parbox[t]{2mm}{\rotatebox[origin=c]{90}{\tiny{\,\,\,\,After}}}      \begin{tabular}{@{}c@{}}\includegraphics[height=\picheight]{#3}\end{tabular}
}
\begin{figure}[t!]
\vskip 0.1in
  \begin{center}
    \begin{subfigure}[t]{1.0\columnwidth}
      \vskip 0.2in
      \taskexample{}{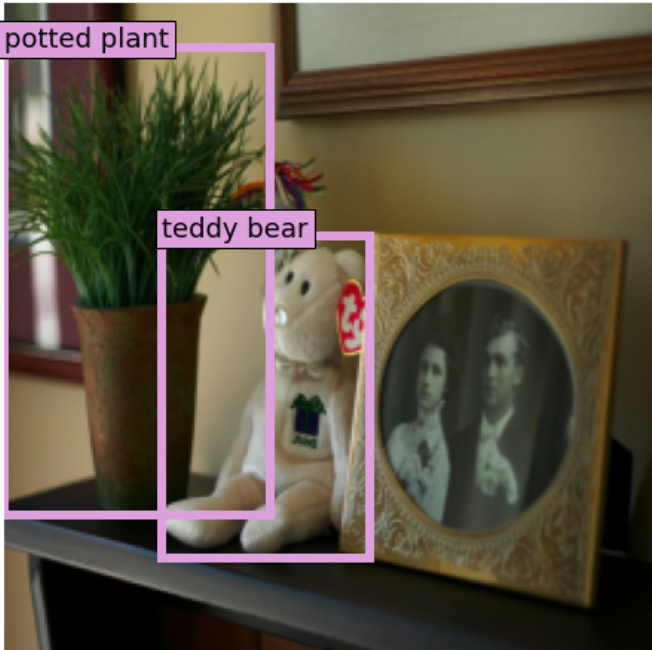}{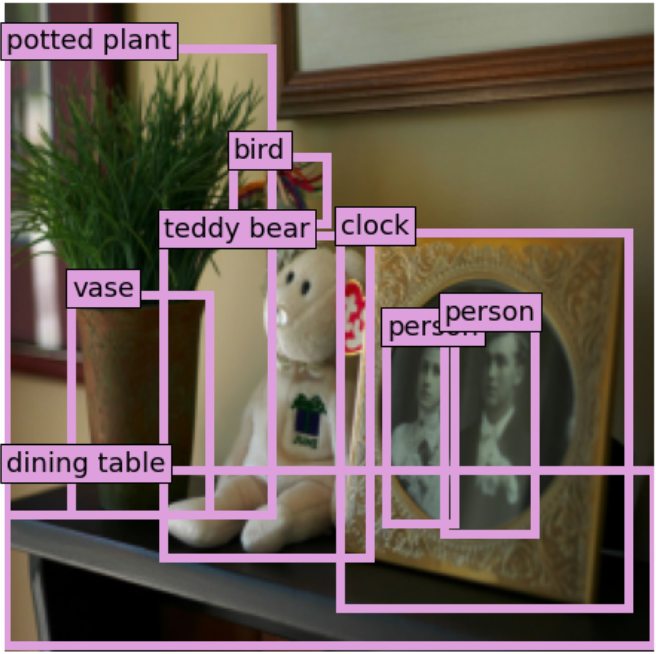}
      \caption{Optimize mAP: $39 \rightarrow 54$, results in a much high recall and learns box prediction confidences.}
      \vskip -0.2in
    \end{subfigure}
    \begin{subfigure}[t]{1.0\columnwidth}
      \vskip 0.2in
      \taskexample{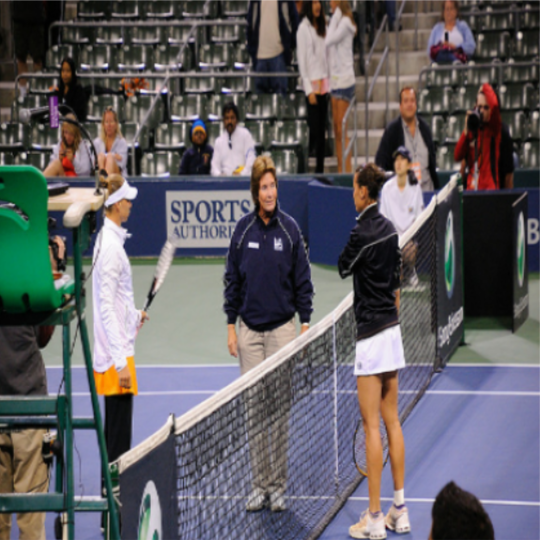}{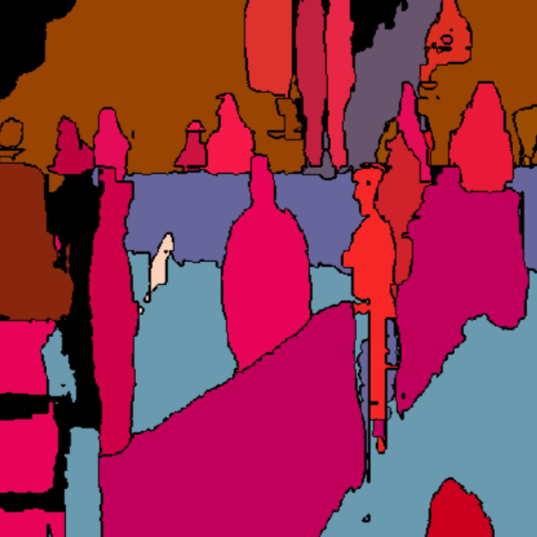}{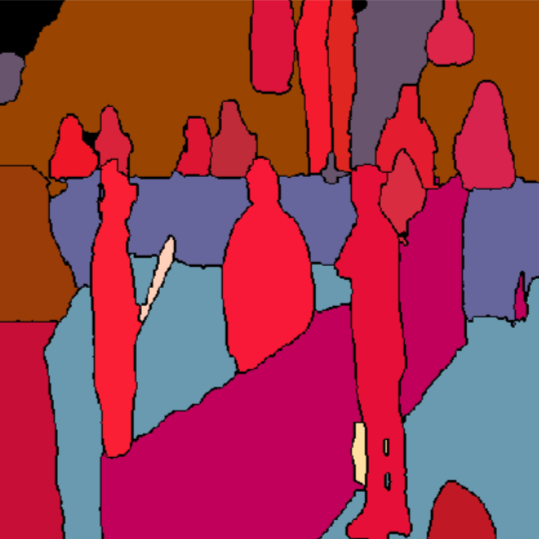}
      \caption{Optimize PQ: $43.1 \rightarrow 46.1$, removes many incoherent predictions, especially for small-scale objects.}
      \vskip -0.2in
    \end{subfigure}
    \begin{subfigure}[t]{1.0\columnwidth}
      \vskip 0.2in
      \taskexample{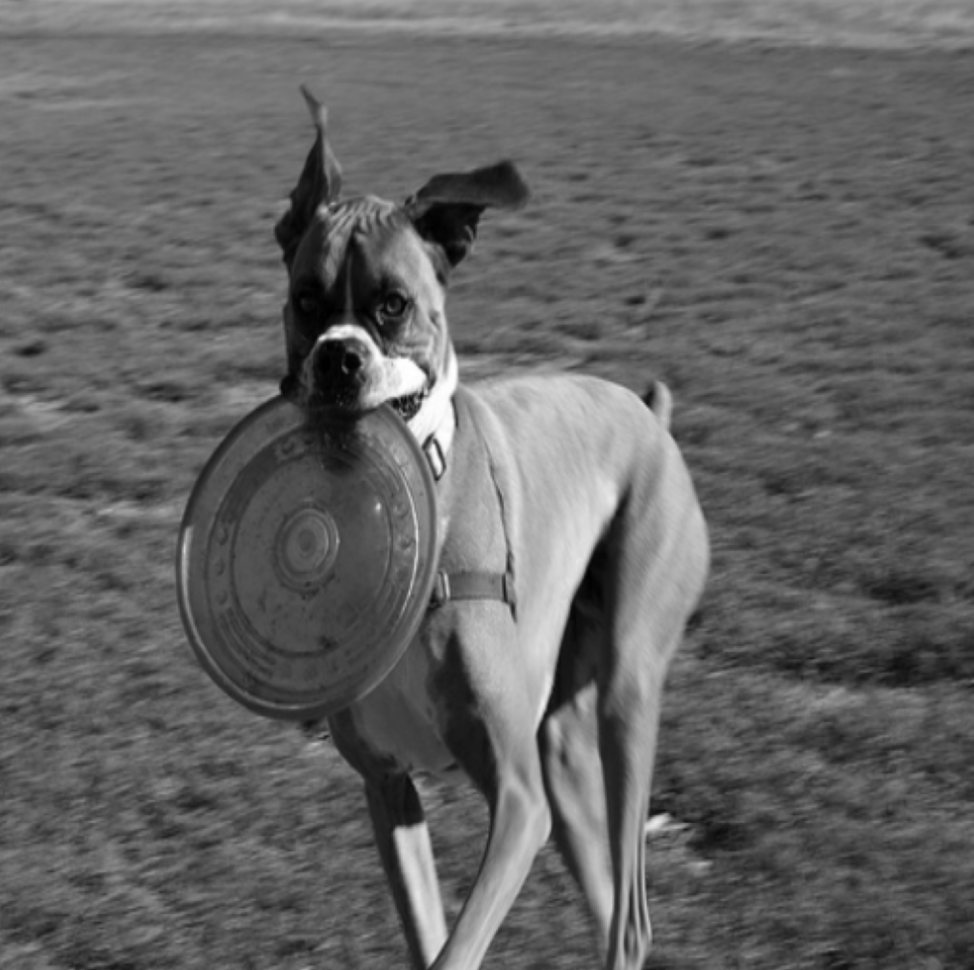}{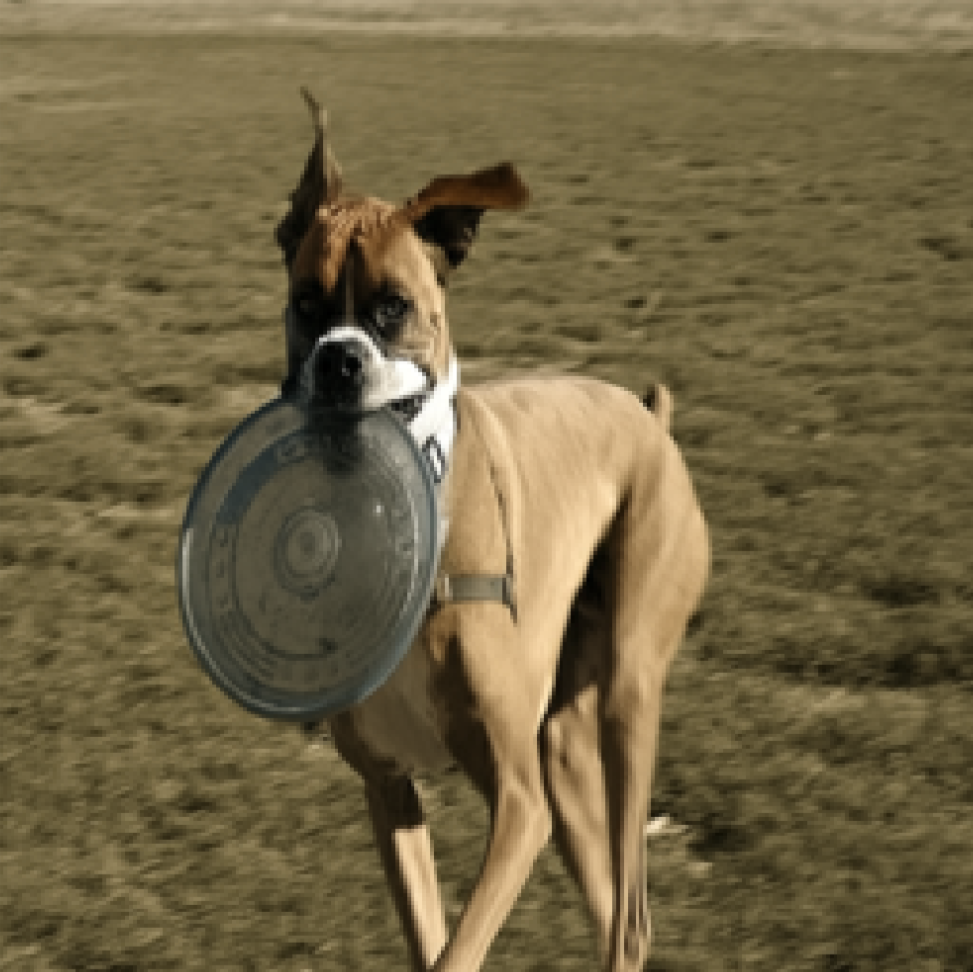}{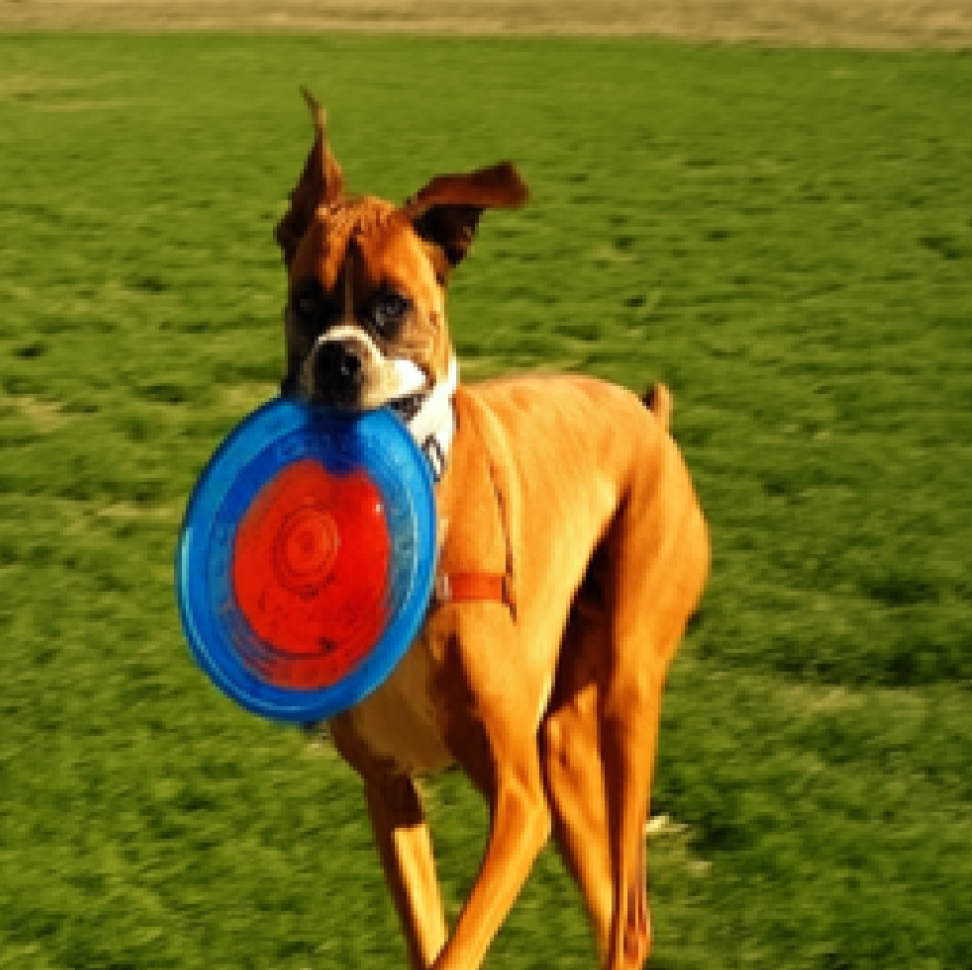}
      \caption{Optimize ``colorfulness'' score: $0.41 \rightarrow 1.79$, improves color diversity and saturation.}
      \vskip -0.2in
    \end{subfigure}

  \caption{By tuning a strong, pretrained model with a reward that relates to the task, we can significantly improve the model's alignment with the intended usage.}
  \label{fig:teaser}
  \end{center}
\vskip -0.2in
\end{figure}

The main criteria for success when dealing with complex outputs in computer vision is not how well the model optimizes the training objective, but rather how well the predictions are aligned with the task risk, i.e.\ the model's performance on the intended usage.
In order to improve this alignment, as a community we iterate on model architectures, data, optimization, sampling procedures, post-processing, etc. As an example, in the context of object detection, researchers use non-maximum suppression post-processing~\cite{ren2015faster,lin2017focal}, set-based global loss~\cite{carion2020end} or even alter the input data~\cite{chen2022pixseq} to obtain models with improved behavior at test time. Although these approaches deliver significant gains, they are often highly specialized to the task and method at hand, while only indirectly optimizing for the task risk.

This problem is not new. It has been extensively studied by the natural language processing (NLP) and reinforcement learning (RL) fields, where it is notoriously hard to formulate an optimization objective for tasks with less tangible goals, such as translation~\citep{kreutzer2018neural} or summarization~\citep{stiennon2020learning}.
A popular approach when dealing with this type of problem is to 
\emph{learn to imitate example outputs, followed by reinforcement-learning to align the model with a reward function}.
Using this approach, the NLP field is now producing exciting results with systems that use large pretrained language models and rewards defined by human feedback to tackle tasks that were otherwise hard to specify~\citep{ouyang2022instructGTP}. Additionally, the same approach is widely adopted for the image captioning task~\citep{rennie2017self}, where \cider{}~\citep{vedantam2015cider} is used as a reward.
Despite that, to the best of our knowledge, reward optimization has not been previously explored for (non-textual) computer vision tasks.

In this work, we demonstrate that tuning a pretrained model with a reward function using \reinforce{}~\cite{williams1992reinforce} works out-of-the-box for a wide range of computer vision tasks.
We illustrate some of our key results in Figure~\ref{fig:teaser}, highlighting both quantitative and qualitative improvements brought by reward optimization for object detection, panoptic segmentation, and image colorization.
The simplicity and effectiveness of our approach on a diverse set of computer vision tasks demonstrates its versatility and adaptability.
Although in this work we mostly use rewards in the form of evaluation metrics, we believe these initial results show promising paths to optimizing computer vision models with more complex and harder to specify rewards, e.g. human feedback or holistic system performance.

\section{Related work}

\textbf{Optimizing computer vision metrics}. 
There is a vast amount of literature in computer vision that sets the goal of optimizing complex non-decomposable or non-differentiable metrics.
In this section we highlight some prominent work. \citet{henderson2017end} propose a specialized approach to compute a pseudo-gradient in order to optimize the average precision (AP) metric for object detection. \citet{song2016training} propose a general framework for computing approximate gradients of metrics.
In the field of semantic image segmentation, CRF loss~\cite{lafferty2001conditional} is often used to ensure segmentation mask consistency. However, the gradient of the CRF-based loss is generally intractable to compute, so many approximations~\cite{krahenbuhl2011efficient} or constrained CRF variants~\cite{nowozin2011decision,kolesnikov2014closed} were proposed in the literature.
In contrast, we propose a generic way to optimize arbitrary rewards that are aligned or coincide with the task risk, for models that are capable of sampling predictions.

One of the most closely related works is \citet{huang2021metricopt}, which proposes an online algorithm that approximates the task reward value with a neural network. This differentiable reward approximation is then used to tune a model. In contrast, we suggest to directly optimize the reward function by relying on the well-known log-derivative trick and the ability of the underlying model to sample multiple predictions. 

\textbf{Optimizing text generation}. 
\citet{ranzato2015sequence} demonstrate improved results on captioning, translation and summarization tasks by training text models with a mixture of MLE and \reinforce{}~\citep{williams1992reinforce} to optimize the non-differentiable rewards (BLEU and ROUGE). \citet{shen2015minimum} also optimizes translation for evaluation metrics but by aproximating the posterior distribution with samples. \citet{rennie2017self} show that using independent model samples as baseline and optimizing \cider{} is simple yet highly effective for image captioning. \citet{keneshloo2019deep} provide a survey of text tasks and uses of RL in seq2seq models. Recently, more advanced RL techniques incorporating human feedback have been used by \citet{ouyang2022instructGTP,glaese2022improving} to align large language models with human intent.

\textbf{Generalization of sampled outputs}: Several works such as \citet{ranzato2015sequence} and \citet{bengio2015scheduled} discuss exposure bias, i.e.\ the distribution discrepancy of previous tokens between training and generation, as a cause for low sample quality. They explore approaches that include sampling from the model during training. \citet{schmidt2019generalization} argues that generalization, and not exposure bias, is the underlying issue to address.
\citet{stahlberg2019nmt} point out that even when using large beams and exact inference, translation models can fail by considering empty and smaller sentences as more likely. Nucleus sampling~\citep{holtzman2019nucleus} was designed to alleviate sampling degeneration.
\citet{leblond2021machine} explore different sampling procedures in translation aided by consistency scores (e.g.\ multilingual BERT).
\citet{ramesh2021dalle} train a model to generate images from text and use a pretrained contrastive model to filter out images inconsistent with the text. \citet{chen2022pixseq} train a generative model for object detection, however good performance of the model is contingent on example augmentation and modified sampling procedures.

\textbf{Reinforcement learning in vision}. Many previously proposed vision models also leverage reinforcement learning algorithms for vision tasks.  They generally focus on learning a system that sequentially attends to various parts of the image and does iterative refinement of the outputs. A prominent example is~\cite{mathe2016reinforcement}, which learns a sequence of image ``glimpses'' that extract visual features from the specific regions and iterative box predictions. See~\cite{le2021deep} for the wide overviews of these type of approaches for object detection and other vision tasks. We largely differ from these approaches, as we do not change the underlying model architecture and instead tune a base vision model to optimize the task-specific reward.

\section{Tuning models with rewards}
Without loss of generality, we formulate a computer vision task as learning a function that maps an input $x$ (in our case an image) to an output represented as a sequence of values $y = [y_1, y_1, \dots, y_n]$ (e.g. sequence of text tokens, sequence of bounding boxes, per-pixel outputs). We assume availability of a dataset of $N$ training examples $D = \{ (x^i, y^i) \}_{i=1}^N$ sampled from the distribution $\mathcal{D}$. When describing algorithms we use bold $\bm{x}$ or $\bm{y}$ to describe a mini-batch of items. Our goal is to learn a conditional distribution $P(y|x,\theta)$ parameterized by $\theta$ that maximizes a reward function $\mathcal{R}$, which coincides or closely aligns with the task risk. Formally, we want to solve the following optimization problem 
\begin{equation}
\max\limits_\theta \E\limits_{x \sim \mathcal{D}} \left[\E\limits_{y \sim P(\cdot|x,\theta)} \mathcal{R}(x, y)\right]   
\end{equation}
 
Our proposed framework for solving the above problem is very simple, consisting of two steps: (1) model pretraining with maximum-likelihood estimation (2) model tuning for the task risk by maximizing a related reward with the \reinforce{} algorithm. We first describe these steps algorithmically and later discuss the intuition and motivation behind the proposed approach.

\begin{figure}[t]
  \centering
  \includegraphics[width=0.85\columnwidth]{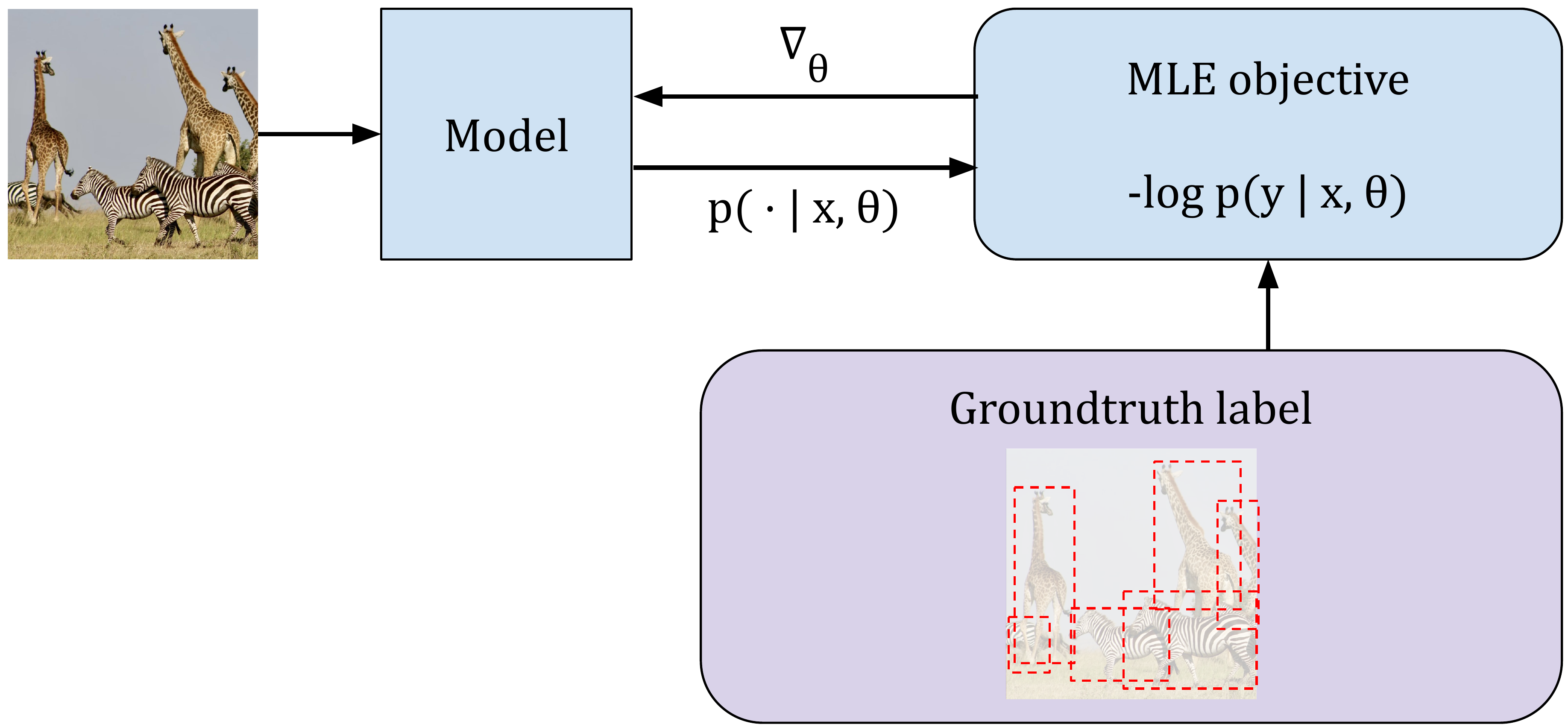}
  \caption{\textbf{Step 1: Maximum-likelihood training.} In a first step, the model is trained to maximize the likelihood of the ground-truth annotations. This is the most common way to train a model and corresponds to learning to imitate the collected data.}
  \label{fig:step1}
\end{figure}

\textbf{Maximum-likelihood pretraining.}
We first use the maximum likelihood principle to estimate parameters $\theta$ and capture the distribution of training data. This can be done with the gradient descent algorithm by maximizing the log-likelihood $\sum\nolimits_{i=1}^{N} \log P(y^i|x^i, \theta)$ of the training data.
Algorithm~\ref{alg:opt_mle} and Figure~\ref{fig:step1} describe the MLE optimization step, which is the most common way of training a model. We refer to the model resulting from this step as the MLE model.

\begin{algorithm}[t]
   \caption{MLE optimization step}
   \label{alg:opt_mle}
\begin{algorithmic}
   \FUNCTION{batch\_loss($\theta$, $\bm{x}$, $\bm{y}$):}
        \STATE {\color{gray}\# $n$ is the size of a mini-batch.}
        \STATE \bfseries{return} $\frac{1}{n} \sum_{i=1}^n \left( \log P(y^i|x^i; \theta) \right)$
   \ENDFUNCTION
   \STATE
   \FUNCTION{step\_mle($\theta$, $\bm{x}$, $\bm{y}$, $\alpha$):}
       \STATE $G_{mle}$ := $\nabla_\theta$ batch\_loss($\theta$, $\bm{x}$, $\bm{y}$)
       \STATE \bfseries{return} $\theta + \alpha G_{mle}$
   \ENDFUNCTION
\end{algorithmic}
\end{algorithm}

\begin{figure}[t]
  \centering
  \includegraphics[width=0.85\columnwidth]{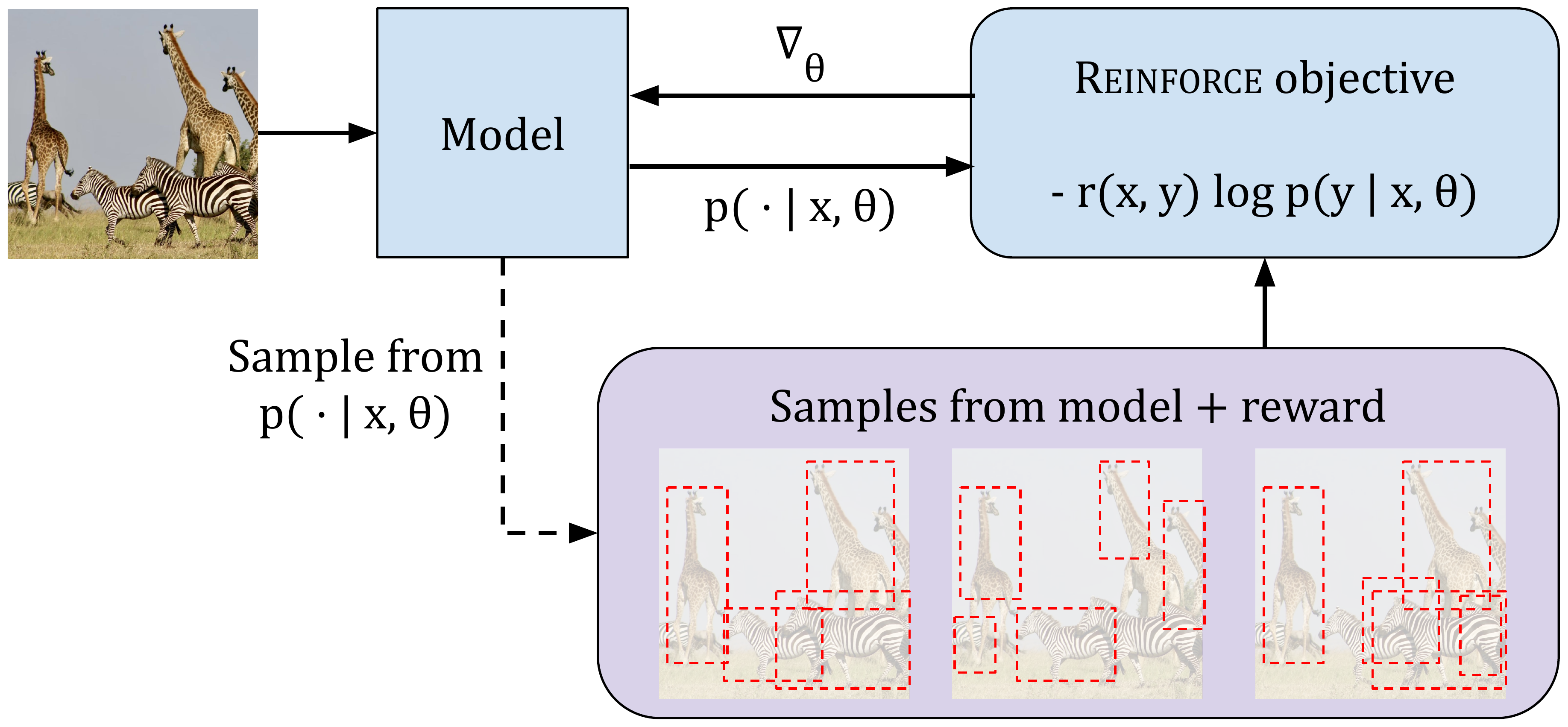}
  \caption{\textbf{Step 2: Reward tuning.} In a second step, the model is further trained to maximize a reward function.
  This is done using \reinforce{} by adjusting the likelihood of model outputs according to their reward.}
  \label{fig:step2}
\end{figure}

\textbf{Reward maximization with \reinforce{}.}
In order to further tune the MLE model to the task risk, we maximize a related reward function. We leverage the \reinforce{} algorithm (also known as the ``log-derivative trick'') to estimate the gradient of the expected reward for a given input $x$:
\[ \nabla_\theta \E_{y \sim P} \left[ \mathcal{R}(x, y) \right] 
= \E_{y \sim P} \left[ \mathcal{R}(x, y) \nabla_\theta \log P(y|x; \theta) \right].
\]
Note that the unbiased estimate of the right-hand side of this equation can be computed as an average of per-example gradients and does not require the reward function to be differentiable. In order to reduce the variance of this gradient estimate, it is common to subtract a baseline value $b$ (independent of the considered example) from the reward function.
In practice, we draw two sample outputs for one training input, use one to estimate the gradient and the other to compute the baseline reward $b$. We provide pseudocode in Algorithm~\ref{alg:opt_reward} and illustrate the procedure in Figure~\ref{fig:step2}.

\begin{algorithm}[t]
   \caption{Reward optimization step}
   \label{alg:opt_reward}
\begin{algorithmic}
   \FUNCTION{batch\_loss($\theta$, $\bm{x}$, $\bm{y}$, $\bm{r}$):}
        \STATE \bfseries{return} $\frac{1}{n} \sum_{i=1}^{n} \left( r \log P(y^i|x^i; \theta) \right)$
   \ENDFUNCTION
   \STATE
   \FUNCTION{step\_reward($\theta$, $\bm{x}$, $\alpha$):}
       \STATE $\bm{y}_{sample}$ := batch\_sample($\theta$, $\bm{x}$)
       \STATE $\bm{y}_{baseline}$ := batch\_sample($\theta$, $\bm{x}$)
       \STATE $\bm{r}$ := $\mathcal{R}$($\bm{x}$, $\bm{y}_{sample}$) - $\mathcal{R}$($\bm{x}$, $\bm{y}_{baseline}$)
       \STATE $G_{r}$ := $\nabla_\theta$ batch\_loss($\theta$, $\bm{x}$, $\bm{y}_{sample}$, $\bm{r}$)
       \STATE \bfseries{return} $\theta + \alpha G_{r}$
   \ENDFUNCTION
\end{algorithmic}
\end{algorithm}

\textbf{Discussion.} The two optimization steps outlined above have complementary strengths and weaknesses.  In practice, neither of them in isolation is sufficient to optimize for a task, but when chained together they work very well.

The first step, model training via the conditional maximum-likelihood estimation, is one the most studied and well-understood approaches in machine learning. There are now very powerful and efficient probabilistic models, e.g. the Transformer encoder-decoder model~\cite{vaswani2017attention}, that can be trained with MLE and can capture very complex data distributions. However, these type of models have a crucial shortcoming. While they can excel at capturing the distribution of training and test data, they are agnostic of the actual task risk and may not perform sufficiently well in their intended usage.

Thus, we leverage the \reinforce{} algorithm to further tune the MLE model to optimize an arbitrary reward function related to the task risk. Crucially, it is sufficient to provide only the numerical value of the reward, without any requirements for the reward functions, such as being differentiable, or being able to run it on a computer (e.g. one can use user feedback as reward).
Note that using \reinforce{} from scratch in the computer vision tasks we explore is most likely unfeasible due to the large output space and reward sparsity.
However, by using a pretrained MLE model, we have a good initial sampling strategy and only need a relatively small number of optimization steps to make quick progress in optimizing the reward function.

\section{Practical applications}
\label{sec:applications}

In this section we show several applications of the described approach to optimize models for vision tasks. In most cases we use an encoder-decoder architecture with a ViT~\cite{dosovitskiy2021vit} encoder to process images and an auto-regressive Transformer decoder to model output distributions. We first pretrain the model using maximum-likelihood estimation then tune it with a task reward. For both steps we use a variant of Adafactor~\citep{shazeer2018adafactor} introduced by \citet{zhai2022scaling} as optimizer and sample greedily at inference time to report results.

It is also important to keep in mind that although in this section we treat existing validation metrics as the task risk, in a real scenario those might differ significantly. In such cases, one might require further validation of the model or iterations on the reward design to guarantee improved performance on the intended usage. Overall our goal is to show that reward optimization is a suitable and general approach to improve computer vision model performance.

\subsection{Panoptic segmentation}
Panoptic segmentation~\cite{kirillov2019panoptic} task can be seen as an aggregation of both instance and semantic segmentation and requires a model to produce a coherent scene segmentation, by assigning a label and instance id to pixels. The metric commonly used in related literature is \emph{Panoptic Quality}~(PQ).
PQ is designed to capture the completeness and detail of predictions and measures. It is computed as a within-class average of mean IoU of matched instances (TP), while penalizing for extra predicted instances (FP) and missed ground truth instances (FN):
\[PQ =  \average_{k \in K} \frac{ \sum_{(p,g) \in TP_k} IoU(p, g) }{|TP_k| + \frac{1}{2}|FP_k| + \frac{1}{2}|FN_k|} \]

\textbf{MLE pretraining.} We use UViM panoptic model~\citep{kolesnikov2022uvim} as our MLE pretraining baseline. UViM adopts a ViT-L/16 encoder for \res{512} resolution and a 24 layers auto-regressive decoder. The decoder output is a $256$ discrete sequence of $4$\,k possible tokens which can then be decoded by UViM stage~I models into a \res{512} panoptic output. The model was trained using MLE on COCO panoptic dataset.

\textbf{Tuning for PQ.} The PQ computation is not decomposable as a sum of per-example rewards. We opt to use a reward which is the sum of matched IoUs and a negative weight $w=0.3$ to unmatched predicted instances in an example:
\[reward(y, k) = \left[ {\textstyle\sum}_{(p, g) \in TP_k} IoU(p, g) \right] - w |FP_k| \]
We use \reinforce{} rule to tune the MLE model for this reward, with a batch size of 128 for 30k steps with constant learning rate $10^{-6}$ after a warmup of $4$\,k steps. We observe that out tuning procedure significantly improves the MLE model (see~table~\ref{tab:panoptic_results}). Our visual inspection suggests that the tuned model is better at avoiding incoherent predictions, especially for the small-scale objects, see~\ref{fig:teaser} as an example.

Note that the task here is quite challenging as we are optimizing a model to sample a discrete sequence with little feedback from a complex reward function. The reward (a scalar) of a model output (a 256-length discrete sequence) is computed by using a neural network to decode the sequence into a \res{512} per-pixel panoptic output which is then compared against the ground truth to approximates a PQ-value per example.

\begin{table}[t]
\caption{Panoptic segmentation results on COCO panoptic validation set after reward optimization.}
\label{tab:panoptic_results}
\vskip 0.15in
\begin{center}
\begin{small}
\begin{sc}
\begin{tabular}{cc}
    \toprule
    Model & PQ (\%) \\
    \midrule
    $\text{UViM}_{512 \times 512}$ $\rightarrow$ \emph{Ours} & $43.1 \rightarrow 46.1$ \\
    $\text{UViM}_{1280 \times 1280}$ & $45.8$ \\
    \bottomrule
\end{tabular}
\end{sc}
\end{small}
\end{center}
\vskip -0.1in
\end{table}

\subsection{Object detection}
\label{sec:detection}
In the object detection task the goal is to predict a tight bounding box for objects (e.g. \textit{chair} or \textit{pen}) present in an input image.
The task is notoriously hard due to the complex nature of the output.
Many different approaches, with unique pros and cons, have been proposed in the past.
One group of techniques~\cite{ren2015faster,lin2017focal} predicts a large redundant collection of boxes and then applies specialized post-processing (non-maximal suppression) at test time. Another approach, proposed by~\citet{carion2020end}, relies on the set-based global loss during training. Finally, Pix2seq~\citep{chen2022pixseq} propose to use a generative model to directly model likelihood of the training data encoded as a sequence of discrete values (discretized box coordinates and semantic class labels).

A shared shortcoming of all these approaches is that they do not offer an explicit way to obtain a model aligned with the task risk and, instead, rely on design choices that implicitly modulate object detection model properties. For example, Faster-RCNN models use a two stage box prediction approach to better balance positive and negative boxes. Similarly, RetinaNet uses the \textit{focal loss} to achieve the same effect. On the other hand, Pix2seq alters the training data by jittering the ground-truth boxes and adding ``fake'' boxes to trick the prediction model into outputting more object bounding boxes. 

Instead, in our experiments, we use detection-specific rewards to optimize a vanilla detection data likelihood model (similar to Pix2seq's base model).
Importantly, this bypasses the need of adopting specialized heuristics to optimize for the standard metrics.
We represent a set of bounding boxes as a discrete sequence by discretizing the coordinates in $1000$ buckets, plus one token for the class label and one token for the per-box prediction confidence.
We use the standard ViT-B/16 as image encoder and 6-layer auto-regressive Transformer decoder (with the same configuration as the ViT-B model). Following our approach we pretrain a MLE model and then tune it with rewards for recall and mAP.

\textbf{MLE pretraining.} Following the standard practice, we pretrain the model on the \textit{Objects365} dataset~\cite{shao2019objects365} and further finetune on the COCO~\cite{lin2014microsoft} dataset.
The model is pretrained on the \textit{Objects365} dataset for 400\,k steps using 256 batch size, with a learning rate of $0.001$ and $0.00005$ weight decay.
We linearly warm up the learning rate for the initial 20\,k steps, and then decay it to zero using a cosine schedule.
We then finetune the model on COCO, using a smaller learning rate $10^{-4}$ without weight decay, for $10$\,k steps. Cosine learning rate schedule with $1$\,k warmup steps is adopted.
We used \res{640} resolution for \textit{Objects365} pretraining and \res{1280} resolution for COCO finetuning.
The resulting MLE model achieves 54.1\% average recall@100 and 40.2 mAP score on COCO.

\textbf{Tuning for recall.}
\emph{Average recall @} $N$ is a popular metric for evaluating object detection models and is expected to correlate with usage in retrieval applications. This metric computes the percentage of object instances in the ground truth that are matched (at a certain IoU threshold) to one of the predicted boxes. At most $N$ predictions per image are allowed. Recall for each IoU threshold and semantic class and is computed independently and then averaged.  Our per-image recall reward is implemented as the count of matched ground-truth boxes minus the number of ``duplicate boxes'' (the boxes that have been matched to an already matched ground-truth box) with $0.3$ multiplier.

We tune our MLE model to optimize the recall reward for $100$\,k steps with the constant learning rate of $10^{-6}$. Table~\ref{tab:detection_results} demonstrates that the resulting model successfully optimizes the average metric, pushing its value from $54.4\%$ to $68.4\%$.  

\textbf{Tuning for mean average precision}
\emph{Mean average precision (mAP)} is a metric based on the area under the precision-recall curve of predicted instances of each class that get matched with a given IoU threshold.
Besides generating a set of predicted boxes, models must also annotate each prediction with a confidence score to rank the items in the curves. These differences encode a different task risk than recall. For example, under this definition a model will be penalized for multiple bounding boxes around one object.

One difficulty is that this metric does not decompose into a sum of per-example rewards. We overcome this by noting that mAP metric is well correlated with recall assuming a prediction model does well at ranking the resulting boxes.  
In order to learn box confidences, we use a supervised loss to learn the expected IoU scores of sampled outputs plus the recall reward defined in the previous section.
We additionally improve the reward by computing its value at various IoU ranges (and averaging them) and by weighting each class based on their frequency observed in the training set.

In Table~\ref{tab:detection_results} we confirm that by optimizing the proposed reward we drastically improve the mAP score of the original MLE model from 39.2\% to 54.3\%.
In Pix2seq~\cite{chen2022pixseq}, the same size ViT-B model with a slightly larger $1333\times1333$ resolution and many heuristics achieves 47.1\%.
The best object detection result reported in Pix2seq is 50.0\%, when using a larger ViT-L backbone.
Our strong ViT-B result clearly demonstrates the promise of the proposed task reward tuning.

\begin{table}[t]
\caption{Object detection results on COCO before and after reward optimization.}
\label{tab:detection_results}
\vskip 0.15in
\begin{center}
\begin{small}
\begin{sc}
\begin{tabular}{ccc}
    \toprule
    Model & mAP (\%) & AR@100 (\%) \\
    \midrule
    \emph{Ours} ($\text{reward}_{\text{mAP}}$)& $39.2 \rightarrow 54.3$ & $54.4 \rightarrow 67.2$ \\
    \emph{Ours} ($\text{reward}_{\text{recall}}$) & N/A & $54.4 \rightarrow 68.4$ \\
    \citet{chen2022pixseq} & $47.1$ & N/A \\
    \bottomrule
\end{tabular}
\end{sc}
\end{small}
\end{center}
\vskip -0.1in
\end{table}

\subsection{Colorization}\label{sec:color}

\begin{figure*}[tbp]
    \centering
    \parbox[c]{1.5mm}{\rotatebox[origin=c]{90}{\tiny{\; Before}}}
    \begin{tabular}{@{}c@{}}\includegraphics[height=4.48em]{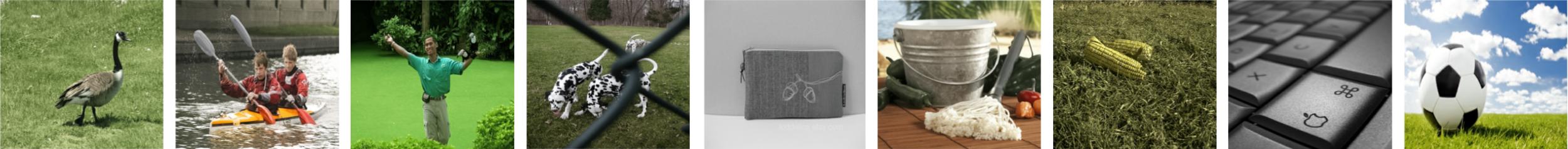}\end{tabular}
    \\
    \parbox[c]{1.5mm}{\rotatebox[origin=c]{90}{\tiny{\quad After}}}
    \begin{tabular}{@{}c@{}}\includegraphics[height=4.5em]{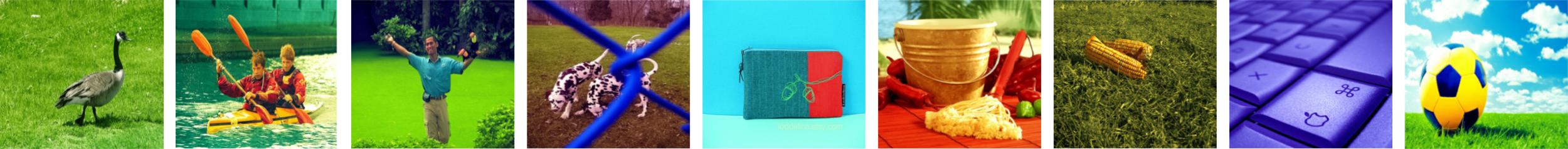}\end{tabular}
    \caption{Random examples demonstrating how UViM colorization model~\cite{kolesnikov2022uvim} predictions change after tuning with a ``colorfulness'' reward. See text in~\ref{sec:color} for the details on the reward function.}
    \label{fig:color}
\end{figure*}

Colorization task is described as adding color to grayscale images. Standard image colorization models are learned by optimizing the likelihood of large datasets of images, i.e. using MLE.
Such model generate plausible image coloring, however often produce faded colors.
In reality, the user of a colorization model may want to produce a vivid image. 
Using our approach we demonstrate that MLE colorization models can be tuned to produce colorful images that are more visually appealing.

\textbf{MLE pretraining.} Similar to the panoptic task, we use UViM colorization as the MLE model. It is a ViT-L/16 encoder for \res{512} resolution images and a $24$ layers auto-regressive decoder. The decoder output is a $256$ discrete sequence which can then be decoded by UViM stage~I models into a \res{512} image. The model was trained using MLE on ImageNet.

\textbf{Tuning for ``colorfulness''.} We design a custom reward that promotes ``colorfulness''. In particular, the reward is a product of two terms that are derived from the input image converted to the \texttt{Lab} colorspace. In this colorspace, the $L$ channel encodes ``lightness'', while the $a$ and $b$ channels encode color. The first term of our reward discourages gray colors. It is defined as the fraction of image pixels that have sufficiently ``vivid color'', where vivid color is defined as $a^2 + b^2 > 10$. The second term of our reward promotes color diversity. It is defined as the image-level entropy of the hue value, with hue being computed by $\arctan(\frac{b}{a})$. Note that to compute the entropy we discretize hue into 7 discrete values, distributing the bins uniformly within the range where hue values are defined.

We tune the MLE model with this reward for $1$\,k steps using a constant learning rate of $3 \cdot 10^{-7}$. As a result of this tuning step, the first reward term grows from $0.46$ to $0.97$, indicating that the vast majority of predicted colors have become more vivid. The second reward term, hue entropy, grows from $1.03$ to $1.84$, indicating much greater diversity of predicted colors. We present qualitative results in Figure~\ref{fig:color} which clearly demonstrate that the new model consistently produces more colorful images.

\subsection{Image captioning}

Image captioning refers to the task of generating textual descriptions for given images.
\cider{}~\citep{vedantam2015cider} is a popular automated metric that measure caption quality based on consensus with a set of human-written reference captions for the image. Specifically, it measures the \ngram~similarity against multiple references captions and takes into account the statistics of the whole dataset such that words that appear more frequently across all captions are given less weight, as they can be considered to be less informative.
As mentioned before, the use of \reinforce{} to optimize a \cider{} reward is an established technique in image captioning~\citep{rennie2017self}.
We include it in this work for completeness.

\textbf{MLE pretraining.} We pretrain an encoder-decoder Transformer model on COCO captions. We initialize the ViT encoder from the ImageNet21k models provided by \citet{steiner2021train}. For the decoder, we randomly initialize a 6-layer auto-regressive decoder. Additionally, we use the BERT~\citep{devlin2018bert} $30$\,k vocabulary to represent text as a discrete sequence of 128 tokens. We pretrain with batch size $256$ for $5$\,k steps using $1$\,k steps linear warmup followed by cosine schedule with learning rate $3 \cdot 10^{-4}$ and $10$x smaller for the encoder parameters. We experiment with two settings: ViT-B/16 and ViT-L/16 both using the same hyper-parameters.

\begin{table}[b]
\vspace{-2em}
\caption{Results on COCO caption on \citet{karpathy2015deep} test split before and after reward optimization.}
\label{tab:caption_results}
\vskip 0.15in
\begin{center}
\begin{small}
\begin{sc}
\begin{tabular}{rr}
    \toprule
     Model & \cider{} \\
    \midrule
    
    \emph{Ours} (ViT-B) & $120.0 \rightarrow 134.5$ \\
    \emph{Ours} (ViT-L) & $121.7 \rightarrow 138.7$ \\
    
    \citet{wang2022end} & $ \text{N/A} \rightarrow 138.2$ \\
    
    \citet{hu2022expansionnet} & $ 128.7 \rightarrow 143.7$ \\

    \bottomrule
\end{tabular}
\end{sc}
\end{small}
\end{center}
\end{table}

\textbf{Tuning for \cider{}.} As previous works we use \cider{} directly as reward using the training set to compute the statistics for the \ngram{} weights. In this case, we use 7 other samples to estimate the reward baseline.
We optimize with batch size $256$ for $10$\,k steps with $1$\,k linear warmup and constant learning rate $10^{-5}$ with $10$x smaller learning rate for the encoder parameters.  For reference we include two recent works \citet{wang2022end,hu2022expansionnet} which also utilize \cider{} optimization and recent architectures.

The results in Table~\ref{tab:caption_results} show that applying the presented approach results in improvements to the MLE model consistent to observations in prior literature, demonstrating the effectiveness of tuning for a specific task risk. 

\section{Analysis}

\subsection{Reward distribution}
Here, we analyse rewards of the models in the image captioning example. We compare direct samples from
the MLE model (before) to the reward tuned model (after).

\begin{figure}[t]
\vskip 0.2in
    \begin{center}
    \begin{subfigure}[t]{1.0\columnwidth}
      \centerline{\includegraphics[width=1.0\columnwidth]{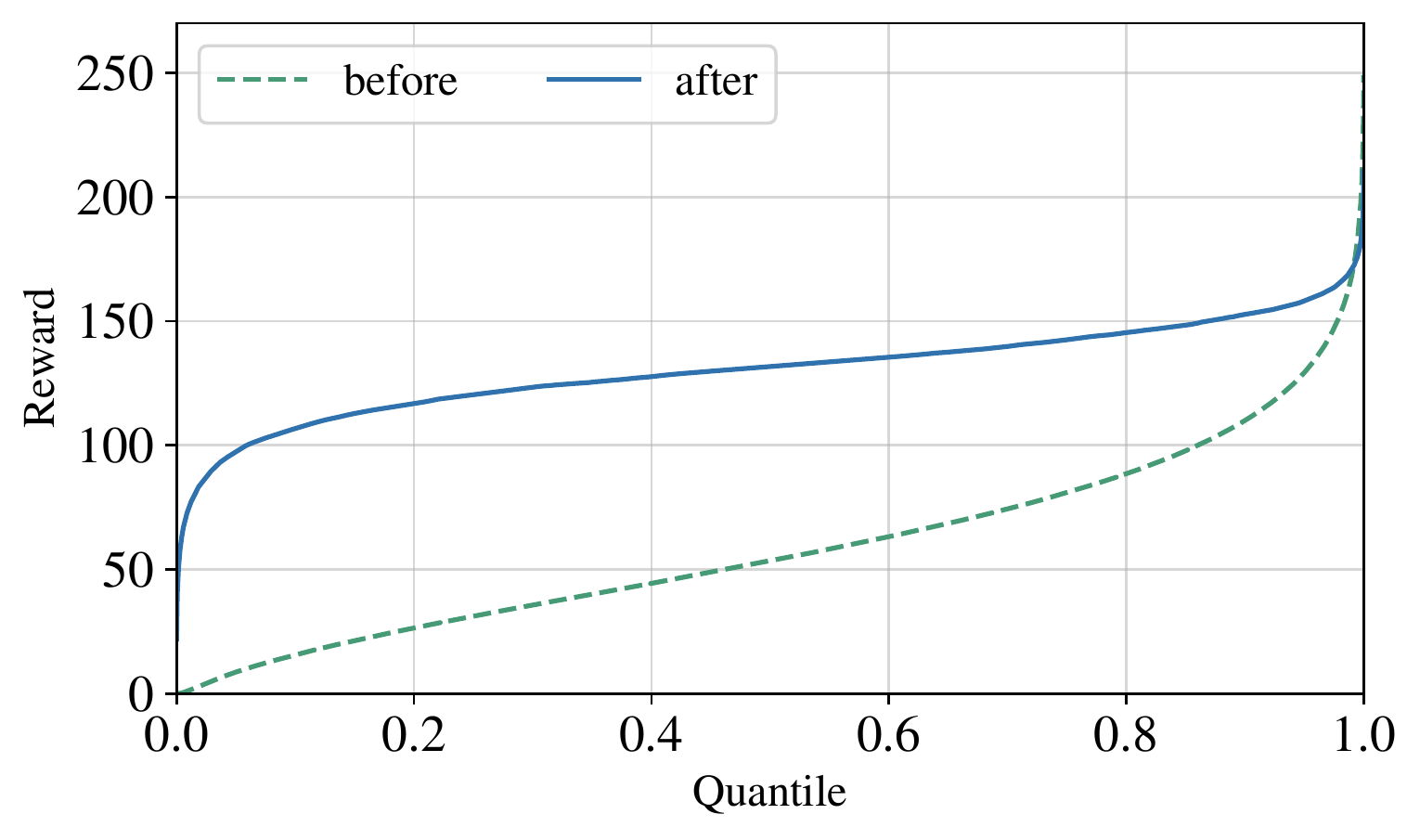}}
      \caption{Quantile function of reward-distribution.}
      \label{fig:cider-reward-dist}
    \end{subfigure}
    \vskip 0.2in
    \begin{subfigure}[t]{.47\columnwidth}
      \includegraphics[width=1.0\columnwidth]{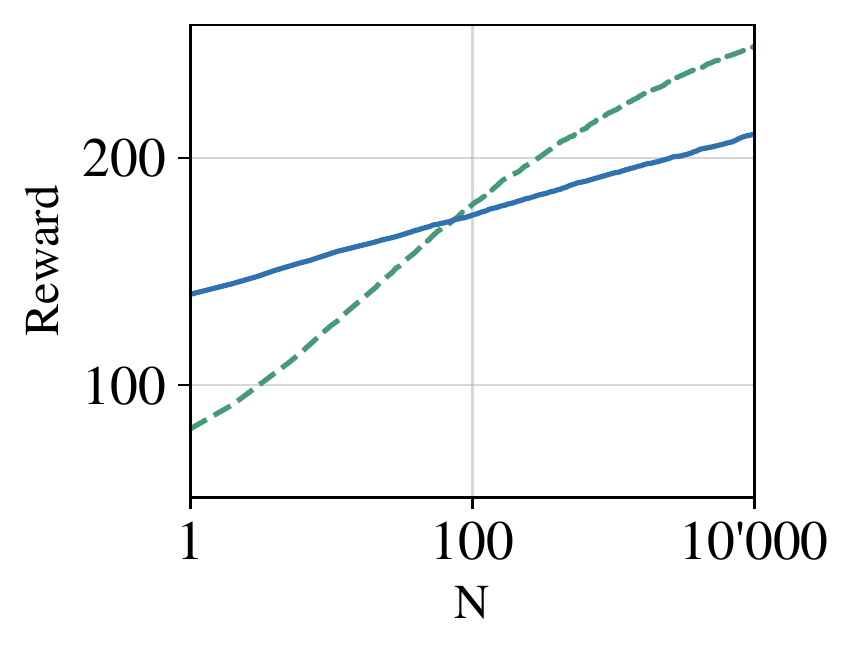}
      \caption{Max reward of N samples.}
      \label{fig:cider-pr-recall}
    \end{subfigure}
    \hfill
    \begin{subfigure}[t]{0.47\columnwidth}
      \includegraphics[width=1.0\columnwidth]{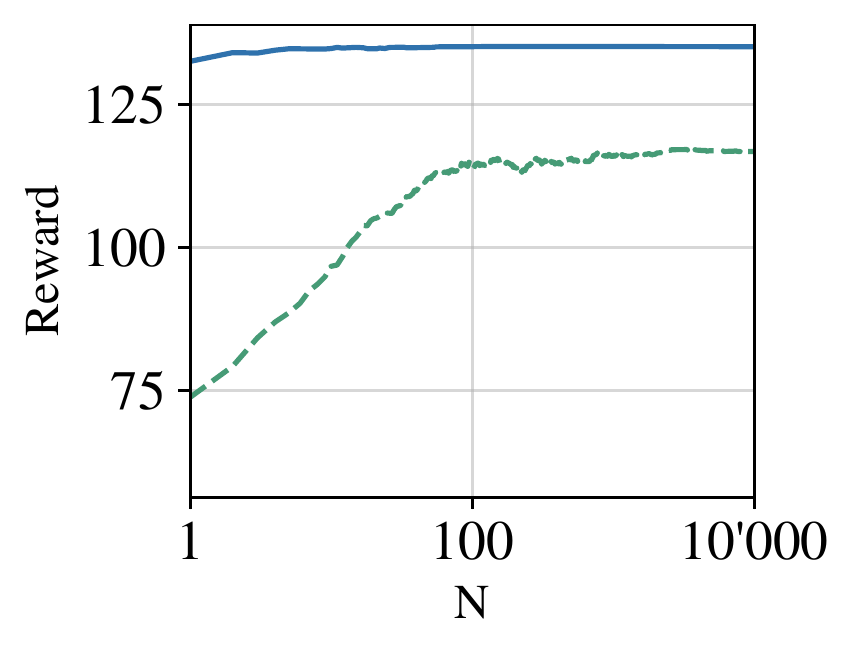}
      \caption{Reward of sample with highest likelihood of N samples.}
      \label{fig:cider-pr-precision}
    \end{subfigure}%
    \caption{Analysis of reward distribution before and after tuning the model for the image captioning task. Measured as mean of 1024 validation examples. In quantile \textbf{plot (a)} the AUC shows the difference between the methods while highlighting that after reward optimization the chance of sampling low-performing samples is greatly reduced. \textbf{Plot (b)} the max reward out of N samples shows that the MLE model includes high-quality outputs in a large enough pool. \textbf{Plot (c)} it is not possible to identify the best outputs using likelihood as observed by the low reward when using the most likely out of N samples even when using $10000$ samples.}
    \label{fig:cider-pr}
    \end{center}
\vskip -0.2in
\end{figure}

We sample $10000$ predictions from each model and plot their rewards using the mean of per-example quantile functions in Figure~\ref{fig:cider-reward-dist}.
The area under the curve of this plot gives the expected reward and it shows the clear improvement of the model: over 50\% of the tuned model samples have an expected reward higher than 125, however for the MLE model only less than 5\% achieves the same standard. Note also, that in the top 1\%-tile, the MLE model is capable of generating very high rewards: this indicates that the MLE model is capable of generating high-quality predictions, however, as we cannot select the best samples at test time, we cannot benefit from them.

To demonstrate this, we additionally illustrate, for each example in the dataset, the reward of (1) the prediction with the highest reward out of $N$ samples in Figure~\ref{fig:cider-pr-recall} and (2) the highest likelihood sample out of $N$ predictions in Figure~\ref{fig:cider-pr-precision}. We aggregate the cross-dataset average of these two statistics and plot them against the number of samples taken for each example N.

Figure~\ref{fig:cider-pr-recall} shows that in a large pool of samples, above 100, there exist better samples in the MLE model pool. However to benefit from this fact, one would need an effective strategy to select the best performing sample.
Figure~\ref{fig:cider-pr-precision} shows what happens when we use max likelihood (e.g. greedy, top-k, nucleus sampling) to select the sample: with the increase of the number of samples $N$, we do see a significant boost in performance for the MLE model inline with greedy/nucleus expectations, however ultimately the performance is much worse than the reward-tuned model even at $N=10000$.

\subsection{Reward-Risk progression}
\begin{figure}[t]
\vskip 0.2in
    \begin{center}
    \begin{subfigure}[t]{1.0\columnwidth}
      \includegraphics[width=1.0\columnwidth]{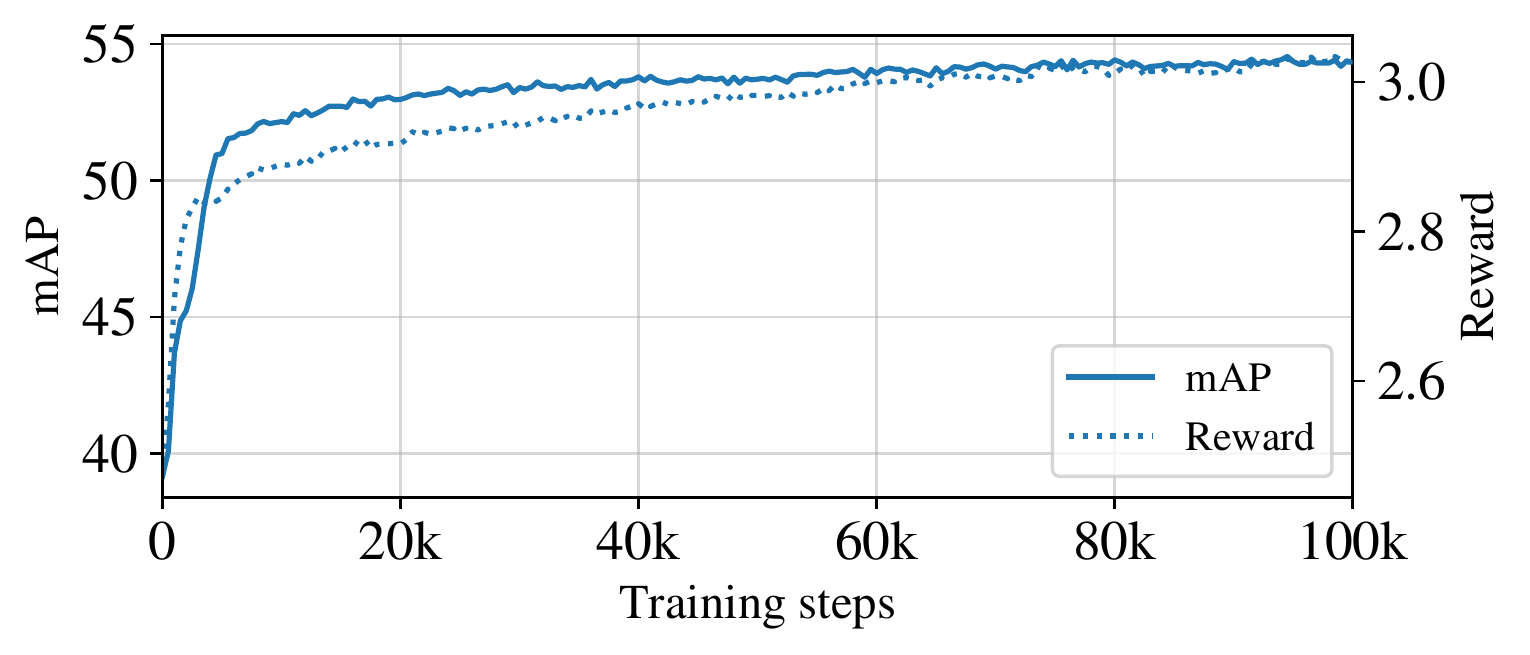}
      \caption{Object detection.}
    \end{subfigure}
    \vskip 0.2in
    \begin{subfigure}[t]{1.0\columnwidth}
      \includegraphics[width=1.0\columnwidth]{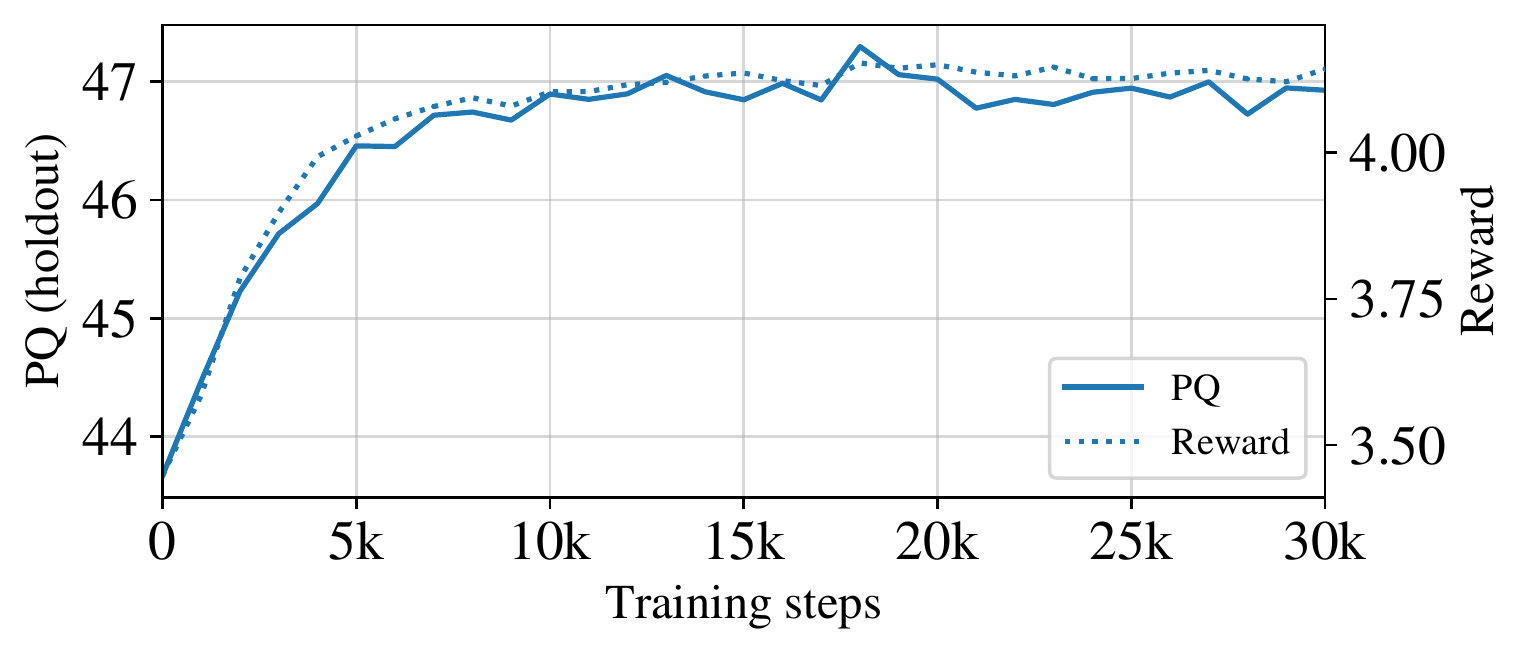}
      \caption{Panoptic segmentation.}
    \end{subfigure}
    \caption{Plot of metrics and rewards measured on the validation set during training for object detection (mAP) and panoptic segmentation (PQ). Overall we observe a good correlation between the reward being optimized and the task risk.}
    \label{fig:train-progression}
    \end{center}
\vskip -0.2in
\end{figure}

In order to translate a task risk into a reward function, often we need to decompose a metric computed for a set of examples into per-example reward.
This could potentially result in undesirable divergence in the progression of per-example reward and the metric.
To see if this is the case empirically,
we plot the progression of reward and goal metrics during training in Figure~\ref{fig:train-progression} for object detection and panoptic segmentation. 
We observe no significant divergence between our reward and the metric.

Additionally, in \cref{fig:train-progression} (a), we observe that the mAP score of object detection quickly goes up for the first $20k$ steps, i.e. from $40.2$\% to $52.3$\%.
With longer reward tuning, the metric keeps going up and it achieves $52.7$\% at $40$\,k steps and $53.2$\% at $60$\,k steps. 

\section{Discussion and Limitations}
\label{sec:discussion}

\textbf{Reward hacking.}
Our work shows the feasibility of tuning a model with rewards beyond standard MLE. However, it is important to note that there is no guarantee that a given reward function will result in improvements in the intended usage.
The model may instead exploit weaknesses in the reward definition, and it is crucial to take that in consideration when validating the models.
In saying that, we think that our demonstrated RL-based approach opens up the possibilities of different forms of rewards, unlocking the potential of using real-risk or human-feedback in training computer vision models, which we believe will lead to great advancements in the current field.

\textbf{Reward design.} In this work we mostly use simple rewards based on existing evaluation metrics. There are however many more options, including combinations of filter-based, input-output checks, simulation checks, use of pretrained models to detect undesired outputs or to keep the model closer to a initial distribution, entropy to encourage diversity and exploration or use of real-world or human feedback.
Another thing to keep in mind is that the reward does not needs to aim to be as exact as the task risk. Alternatives functions may be easier to obtain, control or optimize, for example if the reward provides more guidance than a sparse risk value.
We leave the exploration of this in computer vision tasks to future work.

\textbf{Advanced RL techniques.} The presented approach with warm-up and constant learning rate setup suffices across the explored applications. As such, we saw no need to add regularisation to remain close to the original policy, encourage exploration or attempt to reduce the number of reward calls or model updates. We believe the efficacy is in part due to the MLE-pretrained initialization, allowing the model to avoid potential issues with the simple technique. Although this may not hold in more complex setups, we encourage to try the simple approach in other similar applications.

\textbf{Data for imitation learning.} Can MLE-training alone also reach better alignment with the task goal? This question was part of the motivation of this work. Although we expect more data to help MLE models imitate the ground truth, we found it hard to know what data to collect or how to augment it to have a particular alignment effect with a task risk.
By tuning a model with a reward we obtain that effect by optimizing a model to avoid undesired outputs.
Since the space of undesired outputs where the MLE model assigns high likelihood is hard to predict, it is critical to observe the model in action and focus on the examples the model misassigns high-likelihood.

\textbf{Training cost.} There are two main costs to consider: model sampling cost and the number of queries to the reward function.
Sampling auto-regressive models is notoriously more expensive than computing the likelihood of a given sequence. This is due to difficulties utilizing hardware efficiently and not due to an increase in the number of flops. Note however that this cost is still proportional to inference usage.
Additionally, the presented method only requires a model where the likelihoods of samples can be optimized with gradients. It does not depend on the model being auto-regressive, though that can be an important piece to modelling complex distributions.
For more complex applications the number of queries to the reward function can be a bigger concern. In such cases it is worth to explore off-policy RL techniques and approximate a target reward with a value network.

\section{Conclusion}
Our work shows that reward optimization is a viable option to optimize a variety of computer vision tasks.
Using the simple approach of pretraining to imitate ground truth followed by reward optimization, we were able to:
(a) improve models for object detection and panoptic segmentation trained without other task-specific components
to the level comparable to ones obtained through clever manipulation of data, architectures and losses;
(b) qualitatively affect the results of colorization models to align with a goal of creating vivid and colorful images;
(c) show that the simple approach is competitive with recent works in captioning.

We believe these results demonstrate the possibilities to have more precise control on how models align the non-trivial task risk.
We look forward to more challenging use cases such as tuning scene understanding outputs for robot grasping, where one can optimize the perception models for the probability of a successful grasp.

\section*{Acknowledgements}
We thank Jeremiah Harmsen, Maxim Neumann, Olivier Bachem, Xi Chen and Neil Houlsby for discussions during this project. In addition we thank Jeremiah Harmsen and Matthieu Geist for feedback on the text of this paper.
We use the {\tt big\_vision} codebase \citep{big_vision} for all experiments in this project.



\bibliography{bibliography}

\begin{thebibliography}{43}
\providecommand{\natexlab}[1]{#1}
\providecommand{\url}[1]{\texttt{#1}}
\expandafter\ifx\csname urlstyle\endcsname\relax
  \providecommand{\doi}[1]{doi: #1}\else
  \providecommand{\doi}{doi: \begingroup \urlstyle{rm}\Url}\fi

\bibitem[Bengio et~al.(2015)Bengio, Vinyals, Jaitly, and
  Shazeer]{bengio2015scheduled}
Bengio, S., Vinyals, O., Jaitly, N., and Shazeer, N.
\newblock Scheduled sampling for sequence prediction with recurrent neural
  networks.
\newblock \emph{NeurIPS}, 2015.

\bibitem[Beyer et~al.(2022)Beyer, Zhai, and Kolesnikov]{big_vision}
Beyer, L., Zhai, X., and Kolesnikov, A.
\newblock {Big Vision}.
\newblock \url{https://github.com/google-research/big_vision}, 2022.

\bibitem[Carion et~al.(2020)Carion, Massa, Synnaeve, Usunier, Kirillov, and
  Zagoruyko]{carion2020end}
Carion, N., Massa, F., Synnaeve, G., Usunier, N., Kirillov, A., and Zagoruyko,
  S.
\newblock End-to-end object detection with transformers.
\newblock In \emph{ECCV}, 2020.

\bibitem[Chen et~al.(2022)Chen, Saxena, Li, Fleet, and Hinton]{chen2022pixseq}
Chen, T., Saxena, S., Li, L., Fleet, D.~J., and Hinton, G.
\newblock Pix2seq: A language modeling framework for object detection.
\newblock In \emph{ICLR}, 2022.

\bibitem[Devlin et~al.(2018)Devlin, Chang, Lee, and Toutanova]{devlin2018bert}
Devlin, J., Chang, M.-W., Lee, K., and Toutanova, K.
\newblock Bert: Pre-training of deep bidirectional transformers for language
  understanding.
\newblock \emph{arXiv preprint:1810.04805}, 2018.

\bibitem[Dosovitskiy et~al.(2021)Dosovitskiy, Beyer, Kolesnikov, Weissenborn,
  Zhai, Unterthiner, Dehghani, Minderer, Heigold, Gelly, Uszkoreit, and
  Houlsby]{dosovitskiy2021vit}
Dosovitskiy, A., Beyer, L., Kolesnikov, A., Weissenborn, D., Zhai, X.,
  Unterthiner, T., Dehghani, M., Minderer, M., Heigold, G., Gelly, S.,
  Uszkoreit, J., and Houlsby, N.
\newblock An image is worth 16x16 words: Transformers for image recognition at
  scale.
\newblock In \emph{ICLR}, 2021.

\bibitem[Glaese et~al.(2022)Glaese, McAleese, Trebacz, Aslanides, Firoiu,
  Ewalds, Rauh, Weidinger, Chadwick, Thacker, et~al.]{glaese2022improving}
Glaese, A., McAleese, N., Trebacz, M., Aslanides, J., Firoiu, V., Ewalds, T.,
  Rauh, M., Weidinger, L., Chadwick, M., Thacker, P., et~al.
\newblock Improving alignment of dialogue agents via targeted human judgements.
\newblock \emph{arXiv preprint:2209.14375}, 2022.

\bibitem[Henderson \& Ferrari(2017)Henderson and Ferrari]{henderson2017end}
Henderson, P. and Ferrari, V.
\newblock End-to-end training of object class detectors for mean average
  precision.
\newblock In \emph{ACCV}, 2017.

\bibitem[Holtzman et~al.(2020)Holtzman, Buys, Du, Forbes, and
  Choi]{holtzman2019nucleus}
Holtzman, A., Buys, J., Du, L., Forbes, M., and Choi, Y.
\newblock The curious case of neural text degeneration.
\newblock In \emph{ICLR}, 2020.

\bibitem[Hu et~al.(2022)Hu, Cavicchioli, and Capotondi]{hu2022expansionnet}
Hu, J.~C., Cavicchioli, R., and Capotondi, A.
\newblock Expansionnet v2: Block static expansion in fast end to end training
  for image captioning.
\newblock \emph{arXiv preprint:2208.06551}, 2022.

\bibitem[Huang et~al.(2021)Huang, Zhai, Guo, and Susskind]{huang2021metricopt}
Huang, C., Zhai, S., Guo, P., and Susskind, J.
\newblock Metricopt: Learning to optimize black-box evaluation metrics.
\newblock In \emph{CVPR}, 2021.

\bibitem[Karpathy \& Fei-Fei(2015)Karpathy and Fei-Fei]{karpathy2015deep}
Karpathy, A. and Fei-Fei, L.
\newblock Deep visual-semantic alignments for generating image descriptions.
\newblock In \emph{CVPR}, 2015.

\bibitem[Keneshloo et~al.(2019)Keneshloo, Shi, Ramakrishnan, and
  Reddy]{keneshloo2019deep}
Keneshloo, Y., Shi, T., Ramakrishnan, N., and Reddy, C.~K.
\newblock Deep reinforcement learning for sequence-to-sequence models.
\newblock \emph{IEEE TNNLS}, 31\penalty0 (7):\penalty0 2469--2489, 2019.

\bibitem[Kirillov et~al.(2019)Kirillov, He, Girshick, Rother, and
  Dollar]{kirillov2019panoptic}
Kirillov, A., He, K., Girshick, R., Rother, C., and Dollar, P.
\newblock Panoptic segmentation.
\newblock In \emph{CVPR}, 2019.

\bibitem[Kolesnikov et~al.(2014)Kolesnikov, Guillaumin, Ferrari, and
  Lampert]{kolesnikov2014closed}
Kolesnikov, A., Guillaumin, M., Ferrari, V., and Lampert, C.~H.
\newblock Closed-form training of conditional random fields for large scale
  image segmentation.
\newblock \emph{ECCV}, 2014.

\bibitem[Kolesnikov et~al.(2022)Kolesnikov, Pinto, Beyer, Zhai, Harmsen, and
  Houlsby]{kolesnikov2022uvim}
Kolesnikov, A., Pinto, A.~S., Beyer, L., Zhai, X., Harmsen, J.~J., and Houlsby,
  N.
\newblock {UV}im: A unified modeling approach for vision with learned guiding
  codes.
\newblock In \emph{NeurIPS}, 2022.

\bibitem[Kr{\"a}henb{\"u}hl \& Koltun(2011)Kr{\"a}henb{\"u}hl and
  Koltun]{krahenbuhl2011efficient}
Kr{\"a}henb{\"u}hl, P. and Koltun, V.
\newblock Efficient inference in fully connected crfs with gaussian edge
  potentials.
\newblock \emph{NeurIPS}, 2011.

\bibitem[Kreutzer et~al.(2018)Kreutzer, Khadivi, Matusov, and
  Riezler]{kreutzer2018neural}
Kreutzer, J., Khadivi, S., Matusov, E., and Riezler, S.
\newblock Can neural machine translation be improved with user feedback?
\newblock In \emph{North {A}merican Chapter of the Association for
  Computational Linguistics: Human Language Technologies, (Industry Papers)},
  June 2018.

\bibitem[Lafferty et~al.(2001)Lafferty, McCallum, and
  Pereira]{lafferty2001conditional}
Lafferty, J., McCallum, A., and Pereira, F.~C.
\newblock Conditional random fields: Probabilistic models for segmenting and
  labeling sequence data.
\newblock In \emph{ICML}, 2001.

\bibitem[Le et~al.(2021)Le, Rathour, Yamazaki, Luu, and Savvides]{le2021deep}
Le, N., Rathour, V.~S., Yamazaki, K., Luu, K., and Savvides, M.
\newblock Deep reinforcement learning in computer vision: a comprehensive
  survey.
\newblock \emph{Artificial Intelligence Review}, 2021.

\bibitem[Leblond et~al.(2021)Leblond, Alayrac, Sifre, Pislar, Jean-Baptiste,
  Antonoglou, Simonyan, and Vinyals]{leblond2021machine}
Leblond, R., Alayrac, J.-B., Sifre, L., Pislar, M., Jean-Baptiste, L.,
  Antonoglou, I., Simonyan, K., and Vinyals, O.
\newblock Machine translation decoding beyond beam search.
\newblock In \emph{EMNLP}, 2021.

\bibitem[Lin et~al.(2014)Lin, Maire, Belongie, Hays, Perona, Ramanan,
  Doll{\'a}r, and Zitnick]{lin2014microsoft}
Lin, T.-Y., Maire, M., Belongie, S., Hays, J., Perona, P., Ramanan, D.,
  Doll{\'a}r, P., and Zitnick, C.~L.
\newblock Microsoft coco: Common objects in context.
\newblock In \emph{ECCV}, 2014.

\bibitem[Lin et~al.(2017)Lin, Goyal, Girshick, He, and
  Doll{\'a}r]{lin2017focal}
Lin, T.-Y., Goyal, P., Girshick, R., He, K., and Doll{\'a}r, P.
\newblock Focal loss for dense object detection.
\newblock In \emph{ICCV}, 2017.

\bibitem[Mathe et~al.(2016)Mathe, Pirinen, and
  Sminchisescu]{mathe2016reinforcement}
Mathe, S., Pirinen, A., and Sminchisescu, C.
\newblock Reinforcement learning for visual object detection.
\newblock In \emph{CVPR}, 2016.

\bibitem[Nowozin et~al.(2011)Nowozin, Rother, Bagon, Sharp, Yao, and
  Kohli]{nowozin2011decision}
Nowozin, S., Rother, C., Bagon, S., Sharp, T., Yao, B., and Kohli, P.
\newblock Decision tree fields.
\newblock In \emph{ICCV}, 2011.

\bibitem[Ouyang et~al.(2022)Ouyang, Wu, Jiang, Almeida, Wainwright, Mishkin,
  Zhang, Agarwal, Slama, Gray, Schulman, Hilton, Kelton, Miller, Simens,
  Askell, Welinder, Christiano, Leike, and Lowe]{ouyang2022instructGTP}
Ouyang, L., Wu, J., Jiang, X., Almeida, D., Wainwright, C., Mishkin, P., Zhang,
  C., Agarwal, S., Slama, K., Gray, A., Schulman, J., Hilton, J., Kelton, F.,
  Miller, L., Simens, M., Askell, A., Welinder, P., Christiano, P., Leike, J.,
  and Lowe, R.
\newblock Training language models to follow instructions with human feedback.
\newblock In \emph{NeurIPS}, 2022.

\bibitem[Ramesh et~al.(2021)Ramesh, Pavlov, Goh, Gray, Voss, Radford, Chen, and
  Sutskever]{ramesh2021dalle}
Ramesh, A., Pavlov, M., Goh, G., Gray, S., Voss, C., Radford, A., Chen, M., and
  Sutskever, I.
\newblock Zero-shot text-to-image generation.
\newblock In \emph{ICML}, 2021.

\bibitem[Ranzato et~al.(2015)Ranzato, Chopra, Auli, and
  Zaremba]{ranzato2015sequence}
Ranzato, M., Chopra, S., Auli, M., and Zaremba, W.
\newblock Sequence level training with recurrent neural networks.
\newblock \emph{arXiv preprint:1511.06732}, 2015.

\bibitem[Ren et~al.(2015)Ren, He, Girshick, and Sun]{ren2015faster}
Ren, S., He, K., Girshick, R., and Sun, J.
\newblock Faster r-cnn: Towards real-time object detection with region proposal
  networks.
\newblock \emph{NeurIPS}, 28, 2015.

\bibitem[Rennie et~al.(2017)Rennie, Marcheret, Mroueh, Ross, and
  Goel]{rennie2017self}
Rennie, S.~J., Marcheret, E., Mroueh, Y., Ross, J., and Goel, V.
\newblock Self-critical sequence training for image captioning.
\newblock In \emph{CVPR}, 2017.

\bibitem[Schmidt(2019)]{schmidt2019generalization}
Schmidt, F.
\newblock Generalization in generation: A closer look at exposure bias.
\newblock In \emph{Proceedings of the 3rd Workshop on Neural Generation and
  Translation}, 2019.

\bibitem[Shao et~al.(2019)Shao, Li, Zhang, Peng, Yu, Zhang, Li, and
  Sun]{shao2019objects365}
Shao, S., Li, Z., Zhang, T., Peng, C., Yu, G., Zhang, X., Li, J., and Sun, J.
\newblock Objects365: A large-scale, high-quality dataset for object detection.
\newblock In \emph{ICCV}, 2019.

\bibitem[Shazeer \& Stern(2018)Shazeer and Stern]{shazeer2018adafactor}
Shazeer, N. and Stern, M.
\newblock Adafactor: Adaptive learning rates with sublinear memory cost.
\newblock In \emph{ICML}, 2018.

\bibitem[Shen et~al.(2015)Shen, Cheng, He, He, Wu, Sun, and
  Liu]{shen2015minimum}
Shen, S., Cheng, Y., He, Z., He, W., Wu, H., Sun, M., and Liu, Y.
\newblock Minimum risk training for neural machine translation.
\newblock \emph{arXiv preprint:1512.02433}, 2015.

\bibitem[Song et~al.(2016)Song, Schwing, Urtasun, et~al.]{song2016training}
Song, Y., Schwing, A., Urtasun, R., et~al.
\newblock Training deep neural networks via direct loss minimization.
\newblock In \emph{ICML}, 2016.

\bibitem[Stahlberg \& Byrne(2019)Stahlberg and Byrne]{stahlberg2019nmt}
Stahlberg, F. and Byrne, B.
\newblock On {NMT} search errors and model errors: Cat got your tongue?
\newblock In \emph{EMNLP-IJCNLP}, 2019.

\bibitem[Steiner et~al.(2021)Steiner, Kolesnikov, Zhai, Wightman, Uszkoreit,
  and Beyer]{steiner2021train}
Steiner, A., Kolesnikov, A., Zhai, X., Wightman, R., Uszkoreit, J., and Beyer,
  L.
\newblock How to train your {ViT}? data, augmentation, and regularization in
  vision transformers.
\newblock \emph{arXiv preprint:2106.10270}, 2021.

\bibitem[Stiennon et~al.(2020)Stiennon, Ouyang, Wu, Ziegler, Lowe, Voss,
  Radford, Amodei, and Christiano]{stiennon2020learning}
Stiennon, N., Ouyang, L., Wu, J., Ziegler, D., Lowe, R., Voss, C., Radford, A.,
  Amodei, D., and Christiano, P.~F.
\newblock Learning to summarize with human feedback.
\newblock \emph{NeurIPS}, 2020.

\bibitem[Vaswani et~al.(2017)Vaswani, Shazeer, Parmar, Uszkoreit, Jones, Gomez,
  Kaiser, and Polosukhin]{vaswani2017attention}
Vaswani, A., Shazeer, N., Parmar, N., Uszkoreit, J., Jones, L., Gomez, A.~N.,
  Kaiser, L., and Polosukhin, I.
\newblock Attention is all you need.
\newblock In \emph{NeurIPS}, 2017.

\bibitem[Vedantam et~al.(2015)Vedantam, Lawrence~Zitnick, and
  Parikh]{vedantam2015cider}
Vedantam, R., Lawrence~Zitnick, C., and Parikh, D.
\newblock Cider: Consensus-based image description evaluation.
\newblock In \emph{CVPR}, 2015.

\bibitem[Wang et~al.(2022)Wang, Xu, and Sun]{wang2022end}
Wang, Y., Xu, J., and Sun, Y.
\newblock End-to-end transformer based model for image captioning.
\newblock \emph{arXiv preprint:2203.15350}, 2022.

\bibitem[Williams(1992)]{williams1992reinforce}
Williams, R.~J.
\newblock Simple statistical gradient-following algorithms for connectionist
  reinforcement learning.
\newblock \emph{Machine learning}, 8\penalty0 (3):\penalty0 229--256, 1992.

\bibitem[Zhai et~al.(2022)Zhai, Kolesnikov, Houlsby, and
  Beyer]{zhai2022scaling}
Zhai, X., Kolesnikov, A., Houlsby, N., and Beyer, L.
\newblock Scaling vision transformers.
\newblock In \emph{CVPR}, 2022.

\end{thebibliography}
\bibliographystyle{icml2023}

\clearpage
\appendix

\section{Overview of models and hyper-parameters}

\begin{table}[h!]
\caption{Panoptic segmentation settings.}
\vskip 0.15in
\begin{center}
\begin{small}
\begin{sc}
\begin{tabular}{L{2.5cm}R{3cm}}
    \toprule
    \multicolumn{2}{c}{\emph{Model}} \\
        Encoder: & ViT-L/16 \\
        Decoder: & 24 layers \\
        Seq length: & 256 tokens \\
        Resolution: & \res{512} \\
    \midrule
    \multicolumn{2}{c}{\emph{MLE pretraining}} \\
        Model & \scriptsize{\citet{kolesnikov2022uvim}} \\
    \midrule
    \multicolumn{2}{c}{\emph{Tune for PQ}} \\
        Batch size: & $128$ \\
        Schedule: & constant \\
        Learning-rate: & $1 \cdot 10^{-6}$  \\
        Total steps: & $30~000$ \\
        Warmup steps: & $4~000$ \\
    \bottomrule
\end{tabular}
\end{sc}
\end{small}
\end{center}
\vskip -0.1in
\end{table}

\begin{table}[h!]
\caption{Object detection settings.}
\vskip 0.15in
\begin{center}
\begin{small}
\begin{sc}
\begin{tabular}{L{2.5cm}R{3cm}}
    \toprule
    \multicolumn{2}{c}{\emph{Model}} \\
        Encoder: & ViT-B/16 \\
        Decoder: & 6 layers \\
        Seq length: & 600 tokens \\
        Resolution: & \res{1280} \\
    \midrule
    \multicolumn{2}{c}{\emph{MLE - Objects365 pretraining}} \\
        Resolution: & \res{640} \\
        Batch size: & $256$ \\
        Learning-rate: & $1 \cdot 10^{-3}$ \\
        Weight-decay: & $5 \cdot 10^{-5}$ \\
        Schedule: & cosine  \\
        Total steps: & $400~000$ \\
        Warmup steps: & $20~000$ \\
    \midrule
    \multicolumn{2}{c}{\emph{MLE - COCO finetune}} \\
        Resolution: & \res{1280} \\
        Batch size: & 256 \\
        Learning-rate: & $1 \cdot 10^{-4}$ \\
        Weight-decay: & $0.0$ \\
        Schedule: & cosine  \\
        Total steps: & $10~000$ \\
        Warmup steps: & $1~000$ \\
    \midrule
    \multicolumn{2}{c}{\emph{Tune for recall}} \\
        Batch size: & 256 \\
        Learning-rate: & $1 \cdot 10^{-6}$  \\
        Schedule: & constant \\
        Total steps: & $100~000$ \\
        Warmup steps: & $1~000$ \\
    \midrule
    \multicolumn{2}{c}{\emph{Tune for mAP}} \\
        Batch size: & 256 \\
        Learning-rate: & $1 \cdot 10^{-6}$  \\
        Schedule: & constant \\
        Total steps: & $100~000$ \\
        Warmup steps: & $1~000$ \\
    \bottomrule
\end{tabular}
\end{sc}
\end{small}
\end{center}
\vskip -0.1in
\end{table}

\newpage
\vspace*{0.05in}
\begin{table}[h!]
\caption{Colorization settings.}
\vskip 0.15in
\begin{center}
\begin{small}
\begin{sc}
\begin{tabular}{L{2.5cm}R{3cm}}
    \toprule
    \multicolumn{2}{c}{\emph{Model}} \\
        Encoder: & ViT-L/16 \\
        Decoder: & 24 layers \\
        Seq length: & 256 tokens \\
        Resolution: & \res{512} \\
    \midrule
    \multicolumn{2}{c}{\emph{MLE pretraining}} \\
        Model & \scriptsize{\citet{kolesnikov2022uvim}} \\
    \midrule
    \multicolumn{2}{c}{\emph{Tune for ``colorfulness''}} \\
        Batch size: & $512$ \\
        Schedule: & constant \\
        Learning-rate: & $3 \cdot 10^{-7}$  \\
        Total steps: & $1~000$ \\
        Warmup steps: & $0$ \\
    \bottomrule
\end{tabular}
\end{sc}
\end{small}
\end{center}
\vskip -0.1in
\end{table}

\begin{table}[h!]
\caption{Image captioning settings.}
\vskip 0.15in
\begin{center}
\begin{small}
\begin{sc}
\begin{tabular}{L{2.5cm}R{3cm}}
    \toprule
    \multicolumn{2}{c}{\emph{Model}} \\
        Encoder: & ViT-B/16 / ViT-L/16\\
        Decoder: & 6 layers \\
        Seq length: & 128 tokens \\
        Resolution: & \res{512} \\
    \midrule
    \multicolumn{2}{c}{\emph{MLE pretraining}} \\
        Encoder ckpt: & \scriptsize{ImageNet21k from \citet{steiner2021train}} \\
        Batch size: & 256 \\
        Decoder lr: & $3 \cdot 10^{-4}$ \\
        Encoder lr: & $3 \cdot 10^{-5}$ \\
        Weight-decay: & $5 \cdot 10^{-6}$ \\
        Schedule: & cosine  \\
        Total steps: & $5~000$ \\
        Warmup steps: & $1~000$ \\
    \midrule
    \multicolumn{2}{c}{\emph{Tune for \cider{}}} \\
        Batch size: & 256 \\
        Baseline: & 7 samples \\
        Decoder lr: & $1 \cdot 10^{-5}$  \\
        Encoder lr: & $1 \cdot 10^{-6}$ \\ 
        Schedule: & constant \\
        Total steps: & $10~000$ \\
        Warmup steps: & $1~000$ \\
    \bottomrule
\end{tabular}
\end{sc}
\end{small}
\end{center}
\vskip -0.1in
\end{table}

~
\clearpage

\end{document}